\documentclass[10pt, a4paper]{article}
\usepackage{lrec2022} 
\usepackage{multibib}
\newcites{languageresource}{Language Resources}
\usepackage{graphicx}
\usepackage{tabularx}
\usepackage{soul}
\usepackage{float}
\usepackage{subcaption}
\usepackage{multirow}
\usepackage{titlesec}
\titleformat{\section}{\normalfont\large\bfseries\center}{\thesection.}{1em}{}
\titleformat{\subsection}{\normalfont\SmallTitleFont\bfseries\raggedright}{\thesubsection.}{1em}{}
\titleformat{\subsubsection}{\normalfont\normalsize\bfseries\raggedright}{\thesubsubsection.}{1em}{}
\renewcommand\thesection{\arabic{section}}
\renewcommand\thesubsection{\thesection.\arabic{subsection}}
\renewcommand\thesubsubsection{\thesubsection.\arabic{subsubsection}}

\usepackage{epstopdf}
\usepackage[utf8]{inputenc}

\usepackage{hyperref}
\usepackage{xstring}

\usepackage{color}

\title{The Case for Perspective in Multimodal Datasets}

\name{Marcelo Viridiano\textsuperscript{1}, Tiago Timponi Torrent\textsuperscript{1,2}, Oliver Czulo\textsuperscript{3}, \\ {\bf \large Arthur Lorenzi Almeida\textsuperscript{1}, Ely Edison da Silva Matos\textsuperscript{1}, Frederico Belcavello\textsuperscript{1}}}

\address{\textsuperscript{1} FrameNet Brasil Lab, Graduate Program in Linguistics, Federal University of Juiz de Fora \\
\textsuperscript{2} Brazilian National Council for Scientific and Technological Development – CNPq \\ 
\textsuperscript{3} Institute for Applied Linguistics and Translatology, Universität Leipzig \\
         \{barros.marcelo, arthur.lorenzi\}@estudante.ufjf.br, \{tiago.torrent, ely.matos, fred.belcavello\}@ufjf.br, \\ oliver.czulo@uni-leipzig.de}

\abstract{
This paper argues in favor of the adoption of annotation practices for multimodal datasets that recognize and represent the inherently perspectivized nature of multimodal communication. To support our claim, we present a set of annotation experiments in which FrameNet annotation is applied to the Multi30k and the Flickr 30k Entities datasets. We assess the cosine similarity between the semantic representations derived from the annotation of both pictures and captions for frames. Our findings indicate that: (i) frame semantic similarity between captions of the same picture produced in different languages is sensitive to whether the caption is a translation of another caption or not, and (ii) picture annotation for semantic frames is sensitive to whether the image is annotated in presence of a caption or not.    
 \\ \newline  }

\begin{document}

\maketitleabstract

\section{Introduction}

Multimodal datasets that combine textual and visual information are gaining popularity in Natural Language Processing tasks such as multimodal machine translation \cite{specia-etal-2016-shared,elliott2017findings,elliott-2018-adversarial}, multimodal lexical translation \cite{lala-specia-2018-multimodal}, visual sense disambiguation \cite{gella-etal-2016-unsupervised}, grounded representation of lexical meaning \cite{silberer-lapata-2014-learning}, and lexical entailment detection \cite{kiela-etal-2015-exploiting}. For all of those tasks – and especially for tasks of multimodal translation – the general claim is that using textual data in combination with the ``ground truth'' information provided by a visual mode would improve the performance of multimodal models, allowing them to surpass baselines and equivalent monomodal models. 

In this paper, we describe the first steps into exploring what frame semantic analyses can tell us about multiperspectivity in multimodal datasets annotated in different languages and in different annotation settings. Based on the Flickr 30k dataset \cite{young-etal-2014-image} and its variants – Multi30k \cite{elliott-etal-2016-multi30k} and Flickr 30k Entities \cite{7410660} – we conduct a set of experiments in which automatic frame semantic annotation for image captions and manual frame annotation for images are assessed for their semantic similarity. Image captioning and translation of captions were done by humans. Comparisons adopt both (1) a multilingual perspective with English and Portuguese originals and English-Portuguese translations of image descriptions and (2) multisetup perspective with English image annotations with or without image captions visible to annotators. 

The contributions of this paper are two-fold:

\begin{itemize}
    \item the multilingual setting is a first probing into how similar or different perspectives may be in different languages without digging deeper into how systematic differences might be.
    \item the multisetup perspective tests the assumption that visual information holds some sort of unbiased ``ground truth''.
\end{itemize}

In the remainder of this paper, we discuss, in section \ref{sec:framesemantics}, how Frame Semantics and its computational implementation – FrameNet – incorporate perspective to the core of semantic representations. Next, in section \ref{sec:materials}, we explain compilation, translation and annotation of the corpus used for the experiments devised in section \ref{sec:experiments}. Results and discussion are presented in sections \ref{sec:results} and \ref{sec:discussion}, respectively, while section \ref{sec:conclusion} finalizes the paper.
 
\section{Frame Semantics and Perspective}
\label{sec:framesemantics}
\begin{figure*}[!ht]
\centering
\includegraphics[width=\textwidth]{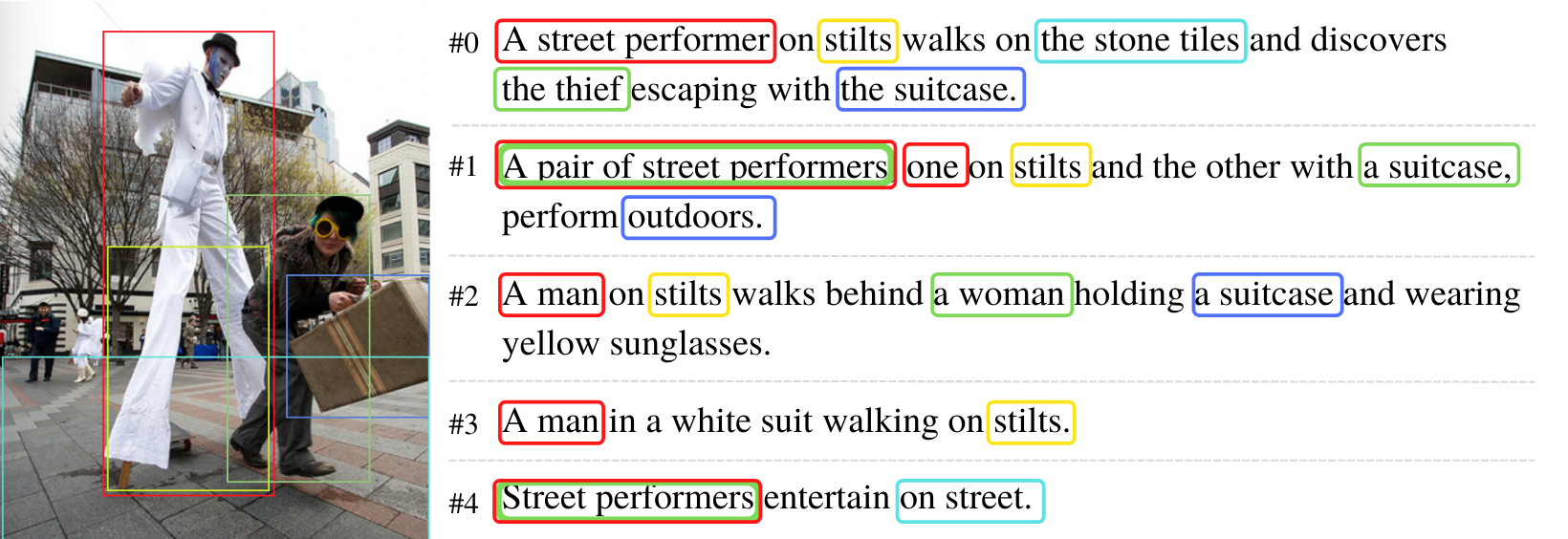}
\caption{\label{fig:entities}Captions and coreference chains from the Flickr 30k Entities dataset.}
\end{figure*}


The main idea behind Fillmore's frame semantics \cite{Fillmore1982} is that human beings understand the meaning of a linguistic expression against the cognitive backdrop of a schematized scene, i.e. a \textit{frame}. A frame is defined as ``any system of concepts related in such a way that to understand any one of them you have to understand the whole structure in which it fits'' \cite[1]{petruck_frame_1996}. 

A central notion in frame semantics is perspective. One of Fillmore's \shortcite{fillmore-85} classic examples refers to the semantic distinction between \textit{coast.n} and \textit{shore.n} in English: Both refer to the \texttt{Relational\_natural\_features} frame, describing the stretch where land mass and sea water meet. While the former lexeme describes the view from the land, the latter is perspectivized from the point of view of the sea.  

Berkeley FrameNet \cite{fillmore_background_2003,johnson2016framenet} was the first lexicographic incarnation of Frame Semantics, building a frame-based lexicon to cover the general vocabulary of English. For each recorded frame that is lexically realizable, the database lists: 
\begin{enumerate}
    \item \textbf{Frame Elements (FE)}: represent the corresponding participants and objects of a frame. Based on the type of supporting information they contribute, frame elements can be categorized as core and non-core. For the frame \texttt{Commerce\_buy}, for instance, \textsc{Buyer} and \textsc{Goods} are core FEs while \textsc{Place} is non-core.
    \item A list of \textbf{Lexical Units}: The list of known lexical units (single or multiple words) that can evoke the frame. For the frame \texttt{Commerce\_buy}, this includes \textit{buyer.n}, \textit{client.n} and \textit{purchase.n}.
    \item \textbf{Frame Relations}: Each frame is connected to related frames with edges denoting the kind of relation (eg. \textit{precedence}, \textit{inheritance}, etc.) that exists between the interconnected frames. \texttt{Commerce\_buy} \textit{inherits} from \texttt{Getting} and is a \textit{perspective on} \texttt{Commerce\_goods-transfer}
\end{enumerate}
Currently, there are FrameNet projects for several languages including German, Japanese and Brazilian Portuguese. A part of these groups forms an initiative for multilingual research in frame semantics, Global FrameNet\footnote{https://www.globalframenet.org/}.

Lately, full-text annotation, translation analysis \cite[i.a.]{czulo_aspects_2017,torrent2018multilingual} and multimodal annotation \cite{belcavello-etal-2020-frame} have seen increasing interest in frame semantics, shifting the annotation focus. In lexicographic annotation, annotation practice usually is aimed at making clear decisions on categories. Annotation accuracy is often measured in relation to a gold standard or by inter-annotator agreement. A lesson from frame-semantic annotation, however, is that not all annotation cases can be decided and multiple interpretations are possible. This issue has seen more attention in the above-mentioned more recent trends. In a phrase such as \textit{Kinder, die dieses Jahr in die Schule gehen} `children who start going to school this year', the annotation of \textit{Schule} `school' in this context could either be argued to be a \textsc{Locale\_by\_use} (a place with a certain purpose) or to refer to \textsc{Education\_teaching} (a domain). In the end, one can wonder whether a decision for either of the two has an added benefit, or whether this actually represents the range of possible interpretations. Indeed, already in the early design of his theory, Fillmore points out that frames (in the terminology at that time `scenes') are associated with and activate each other \cite[124]{fillmore_alternative_1975}. Considering this as well as the fact that frames are, in general, prototypical categories, they mostly cannot be understood as sharply distinct, necessarily discrete categories.

\section{The Framed Multi30k Dataset}
\label{sec:materials}
\begin{figure*}[!ht]
\centering
\includegraphics[width=\textwidth]{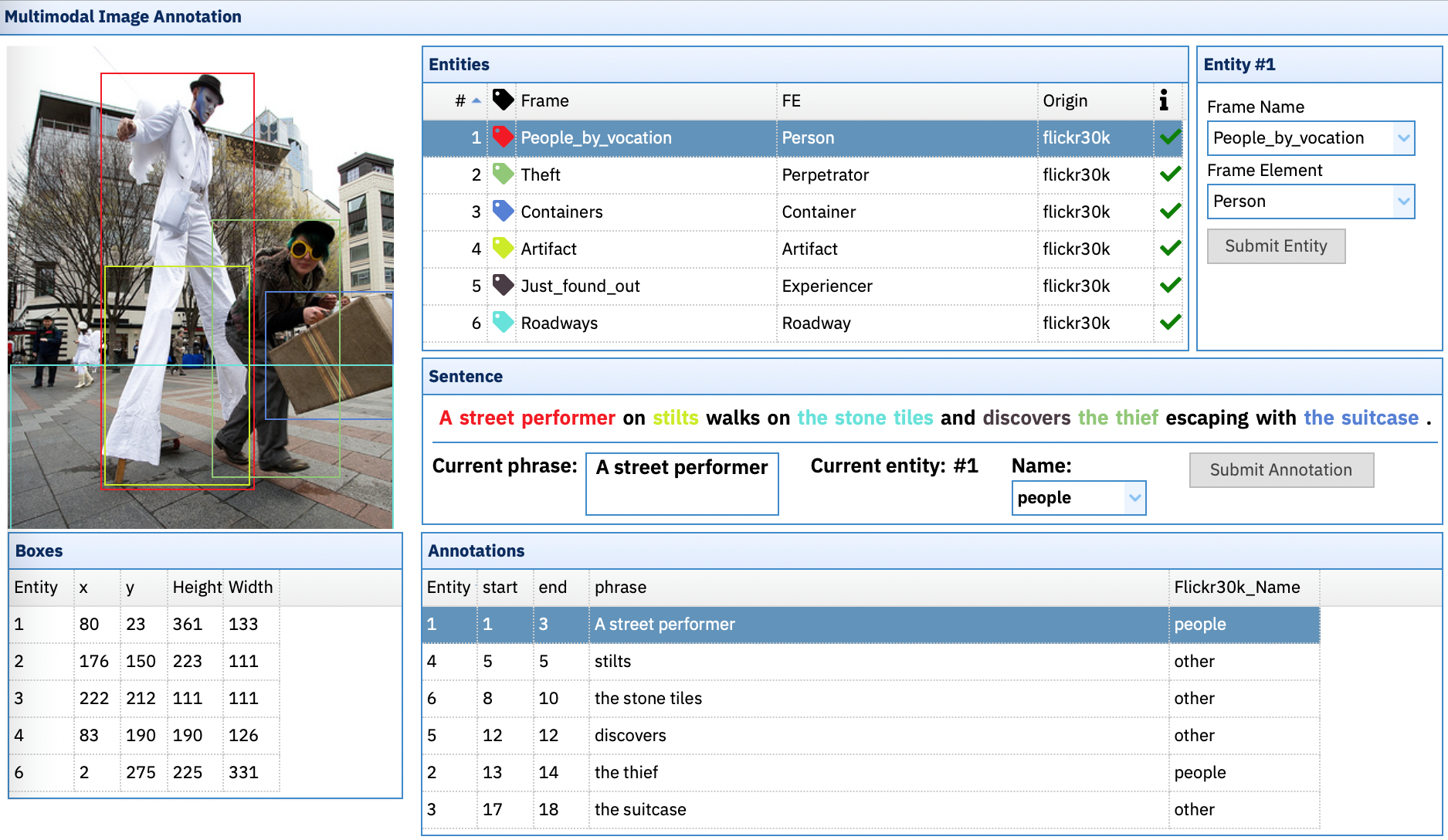}
\caption{\label{fig:caption-anno}User interface for the multimodal annotation tool used for building the Framed Multi30k dataset.}
\end{figure*}

In recent years, several projects have been expanding the popular dataset for sentence-based image description Flickr30k \cite{young-etal-2014-image}: a multimodal dataset containing 31,783 images of everyday activities, events and scenes, each paired with five different English captions providing clear descriptions of the salient entities and events. The Multi30k dataset \cite{elliott-etal-2016-multi30k} is a multilingual expansion of the Flickr30K with five German original conceptual descriptions \cite{hodosh2013framing} crowdsourced independently of the original English captions. The German translations of the English captions were created by professional translators. The Flickr30k Entities dataset \cite{7410660} adds image-to-text relations by manually annotating bounding boxes that assign region-to-phrase correspondences, linking mentions to the same entities in images with lexical items in the five captions describing that image (Figure  \ref{fig:entities}).

For the experiments reported in this paper, we rely on yet another extension of Flickr 30k: the Framed Multi30k dataset \cite{10.3389/fpsyg.2022.838441}, which augments both Flickr30k Entities and Multi30k datasets with (a) semantic annotation based on the network of frames and relations from FrameNet Brasil and (b) Brazilian Portuguese captions to the images. Framed Multi30k includes both English-Portuguese Translation (PTT) of the English Original captions (ENO) and Brazilian Portuguese Original descriptions (PTO) for each image in Flickr 30k. For the translation task, grad students majoring in translation studies were presented with one of the original English captions and instructed to translate the descriptions sticking as closely as possible to the English source sentence. For the task of creating original descriptions, native speakers of Brazilian Portuguese majoring in Language and Linguistics were presented with an image and instructed to write an original conceptual description \cite{hodosh2013framing}. The instruction said to describe only what is depicted in the scene – its entities, their attributes, and relations – as opposed to providing additional background information that cannot be obtained from the image alone, such as about the situation, time, or location in which the image was taken.

As for the annotation of images, Framed Multi30k associates, via manual annotation, the bounding boxes in Flickr 30k Entities with frames and frame elements, using the annotation interface shown in Figure \ref{fig:caption-anno}. Note that, based on the original Flickr 30k Entities dataset organization, the bounding boxes presented for annotation are only the ones grounded in the referential noun phrases found in the caption. This is to say that different captions of one same picture usually have different sets of bounding boxes. 

Because Framed Multi30k is still under construction, for this paper, we selected a random sample of 2,000 images for which there already are both the PTT and the PTO captions, in addition to the ENO captions. Those images and their corresponding captions comprise the dataset used in the experiments.

Portuguese original captions are usually shorter than the original English captions, both in terms of the number of tokens (27,421 PTO x 38,881 ENO), types (3,747 PTO x 3,809 ENO), and characters (128,659 PTO x 152,837 ENO). When comparing the original descriptions with their translations to Brazilian Portuguese, we observe that, despite having fewer tokens (35,074 PTT x 38,881 ENO), translated sentences are more varied in terms of types (4,712 PTT x 3,809 ENO) and have a higher character count (165,724 PTT x 152,837 ENO). The properties of PTO and PTT captions are similar to the ones found for German original and translated captions when the Multi30k dataset was built \cite{elliott-etal-2016-multi30k}. Inflectional properties of German, in comparison to English, may in part be responsible for the differences; they may also reflect the oft-made observation that translations have a tendency to be longer than originals.

\section{Experiments}
\label{sec:experiments}
\begin{figure*}[!ht]
\centering
\includegraphics[width=\textwidth]{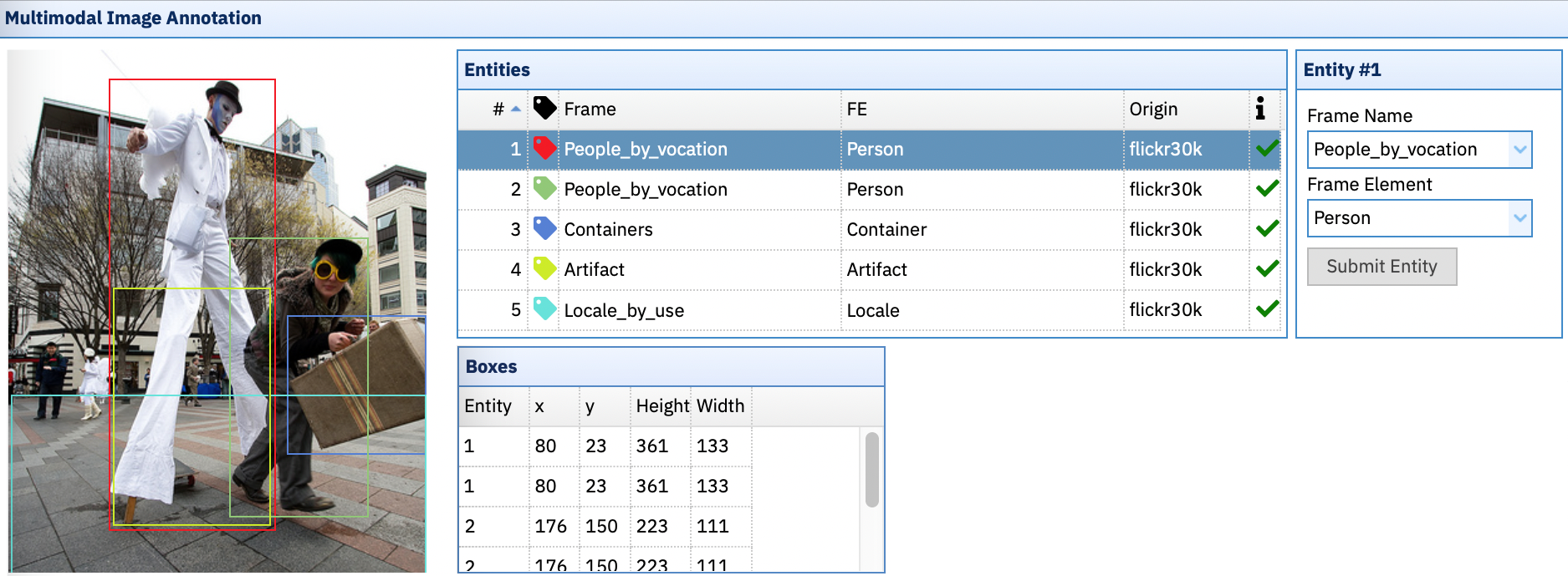}
\caption{\label{fig:vis_anno}User interface of the multimodal annotation tool adapted for the second task.}
\end{figure*}

Experiments were set up so as to assess frame semantic similarity across languages and across communicative modes. In both cases, we use the cosine similarity (CS) between the frame semantic representations generated either automatically by a semantic parser or manually, via annotation.

The CS algorithm used to measure frame semantic similarity between two annotations comprises three stages:

\begin{enumerate}
\item building an associate table using frames from FrameNet;
\item building associative arrays associating each annotation – automatic or manual – with the directly evoked and related frames in the FrameNet database;
\item measuring cosine similarity between associative arrays.
\end{enumerate}

In this application, the FrameNet network is represented as an Acyclic Directed Graph – the FN graph – where each node (frame) is associated with another node (frame) higher in the same hierarchy. The FN graph is used to build the associative table for each frame. This table indicates a relatedness metric and is calculated using Spread Activation (SA) \cite{gouws-etal-2010-measuring}. The SA algorithm models an iterative energy propagation process from one or more nodes to other nodes in a graph in three stages: (i) pre-adjustment, (ii) spreading, and (iii) post-adjustment \cite{crestani1997application}. 
	
Before the spreading stage, the energy value for each node was calculated during the pre-adjustment stage. Energy decay was calculated for the value of the node so that this value is within the [0,1] interval. The calculated value was then output to the neighboring nodes. Post-adjustment was not used, since the FN graph is acyclic and the FN hierarchies do not comprise many levels.

From the associative table of each frame present in an annotation, the associative arrays were built. Each index in the array corresponds to an associated frame and the value of the index indicates the activation level of that frame. When comparing two annotations, for example a1 and a2, the array for a1 is completed with the frames evoked by a2 but not evoked by a1, and vice versa. A zero value is computed for each of those ``completion frames''. Finally, the relatedness between the two annotations is measured using the standard CS between two associative arrays. 

CS of associative arrays were used as a metric for assessing semantic similarity of both captions in a crosslinguistic perspective, and communicative modes in a crossmodal perspective. We describe the experimental designs used for each perspective next.

\subsection{Semantic similarity across languages}
\label{sec:daisy}

To evaluate semantic similarity of image captions across languages, we compare the semantic frame representation of the original English captions (ENO) provided by the Flickr 30K dataset with the Brazilian Portuguese translations of those captions (PTT) and the original Portuguese captions produced for the same pictures (PTO). Comparisons are pairwise and are assessed using the cosine similarity between associative arrays generated for each caption. Therefore, three measures are extracted from this experiment, namely, cosine similarities for the ENO x PTT, ENO x PTO and PPT x PTO pairs. 

Frame-evoking lemmas from all captions in each language were automatically retrieved using DAISY \cite{10.3389/fpsyg.2022.838441}, a disambiguation algorithm that uses the network of semantic frames, Frame Elements (FEs) and Lexical Units (LUs) from the FrameNet database to assign the correct frames to each lexical item based on the context provided by each sentence. By treating word forms, lexemes, LUs and frames as nodes in a graph, and attributing values to each node and also to each candidate lemma, the disambiguation algorithm uses a spread activation search method \cite{DIEDERICH+1990+25+64,tsatsaronis2007word} to calculate the energy decay of each lemma as it propagates through the nodes in the network. This method takes into account not only how far a specific node is from the beginning nodes – meaning, how much energy is lost as connections need to be traversed in order to reach that specific node – but also how it is activated by liked neighboring nodes, receiving more energy, which helps determine its relative importance in the network. 

Table \ref{table:framelemma} presents the differences among the multilingual corpora. Brazilian Portuguese translations have a lower average number of lemmas than the original English captions (19.44 ENO x 17.54 PTT) and also fewer frame-evoking lemmas (17.78 ENO x 12.38 PTT). Original descriptions in Portuguese also have a lower average number of lemmas than the English captions (19.44 ENO vs. 13.71 PTO lemmas) and also evoked fewer frames (17.78 ENO vs. 9.98 PTO frames). Considering the normalized number of frames per lemma, the original English caption ratios (\textit{M} = 0.92, \textit{SD} = 0.33) are significantly higher than the Portuguese translations (\textit{M} = 0.71, \textit{SD} = 0.28) and original descriptions (\textit{M} = 0.75, \textit{SD} = 0.32), with $\textit{t}(3996) = 21.28, \textit{p} < 0.001$ and $\textit{t}(3996) = 16.35, \textit{p} < 0.001$, respectively, using Welch's t-test \cite{welch1947generalization}. This difference is a consequence of the broader lexical coverage of English in relation to Portuguese. It is worth noting, however, that this difference does not impact the validity of the comparisons between cosine similarities because the vectors representing frames evoked by a sample in English are always paired with samples of the other corpora. Any variation in a comparison between a translation and a original sentence in Portuguese is caused by their own differences.

Portuguese originals also have a higher frame:lemma ratio, although this difference is lower than the others ($\textit{t}(3996) = -3.83, p < 0.001$). In this case, since both corpora are on the same language, the difference can be explained by the lexical choices of the annotators: the translation corpus contains 2,755 singletons, while the original Portuguese annotations have 2,118.

\begin{table}[!ht]
\centering
\def\arraystretch{1.4}
\begin{tabular}{ccccc}
\hline
\multicolumn{1}{l}{}           & \multicolumn{1}{l}{} & \multicolumn{1}{l}{ENO} & \multicolumn{1}{l}{PTT} & \multicolumn{1}{l}{PTO} \\ \hline
\multirow{2}{*}{Frames}        & avg. \#              & 17.78                   & 12.38                   & 9.98                    \\
                               & stdev                & 8.04                    & 5.96                    & 4.86                    \\ \hline
\multirow{2}{*}{Lemmas}        & avg. \#              & 19.44                   & 17.54                   & 13.71                   \\
                               & stdev                & 6.47                    & 6.21                    & 5.48                    \\ \hline
\multirow{2}{*}{Frame:Lemma} & avg.                 & 0.92                    & 0.71                    & 0.75                    \\
                               & stdev                & 0.33                    & 0.28                    & 0.32                    \\ \hline
\end{tabular}
\caption{Counts and ratios for annotated frames and for lemmas \label{table:framelemma}}
\end{table}

\subsection{Semantic similarity across modes in different annotation setups}

For assessing similarity of semantic annotation of different communicative modes and how it may be influenced by the annotation setup, the results of two annotation tasks for tagging bounding boxes in the Flickr 30k Entities dataset for frames and frame elements were used. 

The first is the one originally designed for building the Framed Multi30k dataset, where native speakers of Brazilian Portuguese, who are also fluent in English, are assigned the task of enriching the multimodal dataset with FrameNet frame and frame element tags while analyzing image-caption pairings – see Figure \ref{fig:caption-anno}.

Before being assigned this task, annotators were trained on the guidelines and the quality of their annotation work was manually checked for a first batch comprised of six hundred annotations, one hundred images from each of the six annotators involved. After validating the control quality batch, a second batch with 275 images per annotator was assigned to them. For this first task, annotators were instructed to follow the FrameNet annotation guidelines and to assign a semantic frame and a frame element to the objects in the bounding boxes. 

In the example annotation (Fig. \ref{fig:caption-anno}), the lexical items ``A street performer,'' ``stilts,'' ``the stone tiles,'' ``discovers,'' ``the thief,'' and ``the suitcase,'' from the original English caption created to describe this image, are correlated to bounding boxes containing the entities referred to by those lexical items. For the first bounding box – colored red and correlated with the also red sentence segment 
``A street performer,'' – the annotator assigned the frame \texttt{People\_by\_vocation}, which contains words for individuals as viewed in terms of their vocation, and the core frame element \textsc{Person}. For the second bounding box – colored green and correlated with the green sentence segment ``the thief'' – the annotator assigned the frame \texttt{Theft}, which is evoked by lexical items describing situations in which a perpetrator takes goods from a victim, and the core frame element \textsc{Perpetrator} – the person (or other agent) that takes the goods away.

The second annotation task was specifically devised for this paper. We asked a different group of five annotators to annotate a subset of 1,000 images within the original 2,000 images sample for frames and frame elements in the bounding boxes associated with each image. This time, however, annotators were not presented with the captions. In other words, they should apply FrameNet labels while having only the visual mode as reference. Given the nature of the visual corpus, annotators were instructed to only assign Entity frames and the related core frame element. The same manually created bounding boxes from Flickr 30k Entities were used to determine which objects from each image should be annotated. This time, however, all bounding boxes associated with each image were presented (Fig. \ref{fig:vis_anno}). As in the first task, the quality of the annotations was assessed by manually checking a subset before allowing annotators to proceed to the full task.

\section{Results}
\label{sec:results}

In the following subsections we present the results for each of the experiments described in section \ref{sec:experiments}.

\subsection{Similarity of semantic representations across languages}

\begin{figure}[!ht]
    \centering
    \begin{subfigure}[t]{0.48\linewidth}
        \centering
        \includegraphics[width=\linewidth]{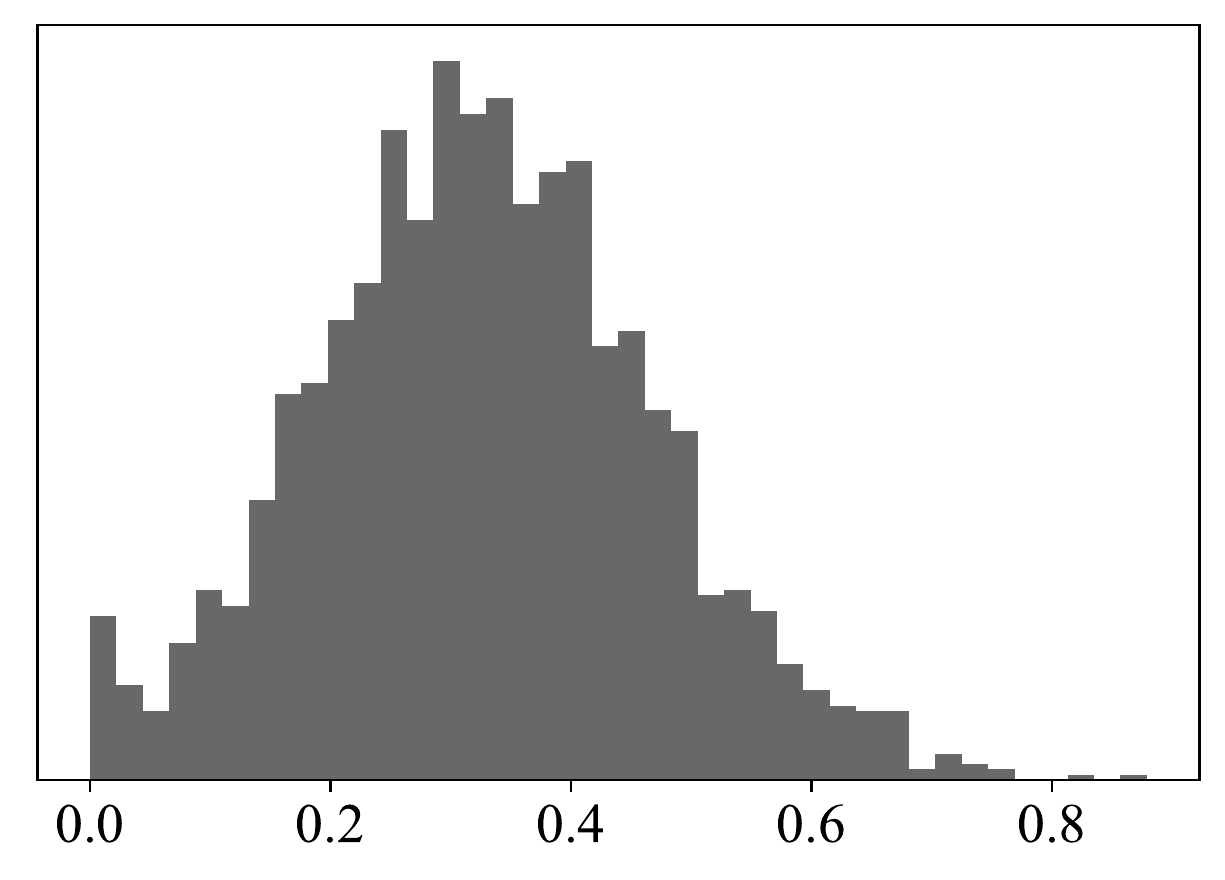}
        \caption{ENO $\times$ PTO}
    \end{subfigure}
    ~
    \begin{subfigure}[t]{0.48\linewidth}
        \centering
        \includegraphics[width=\linewidth]{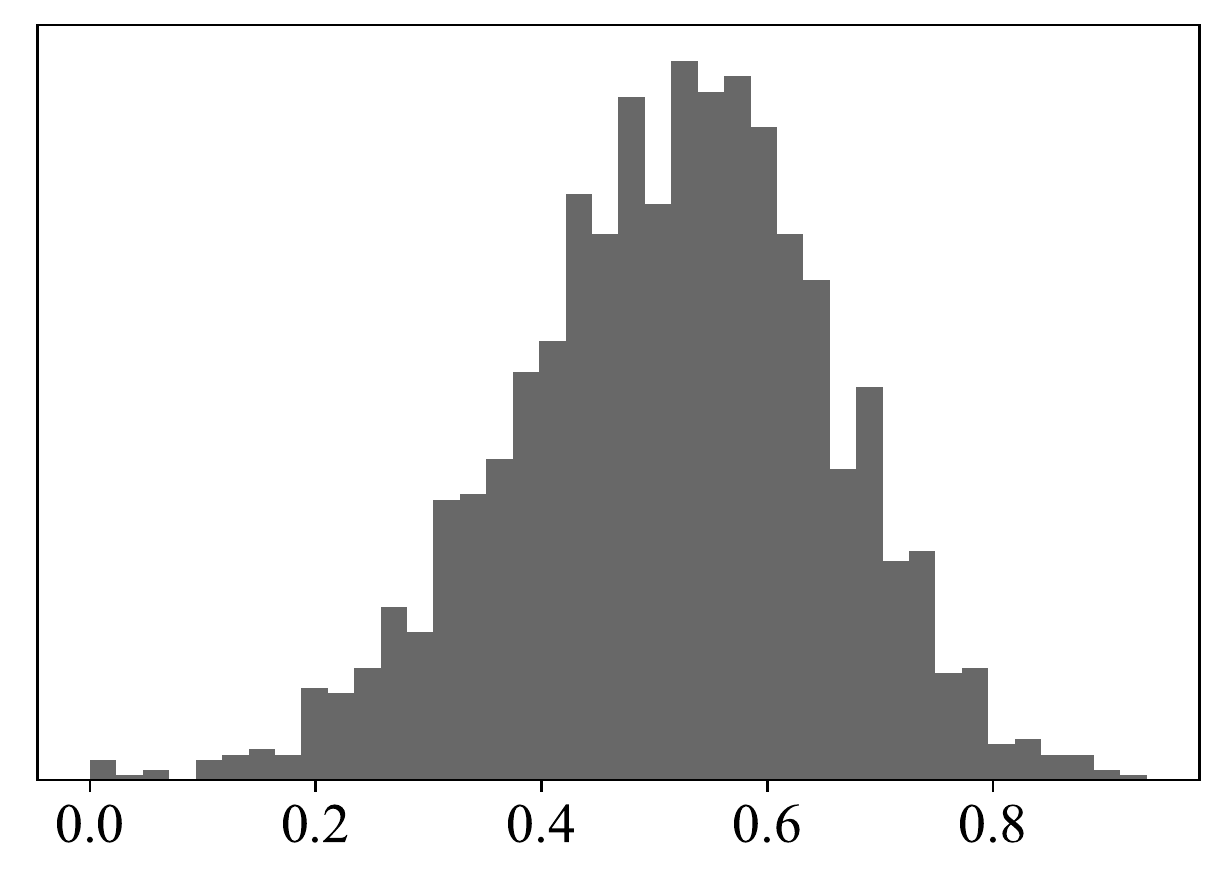}
        \caption{ENO $\times$ PTT}
    \end{subfigure}
    ~
    \begin{subfigure}[t]{0.48\linewidth}
        \centering
        \includegraphics[width=\linewidth]{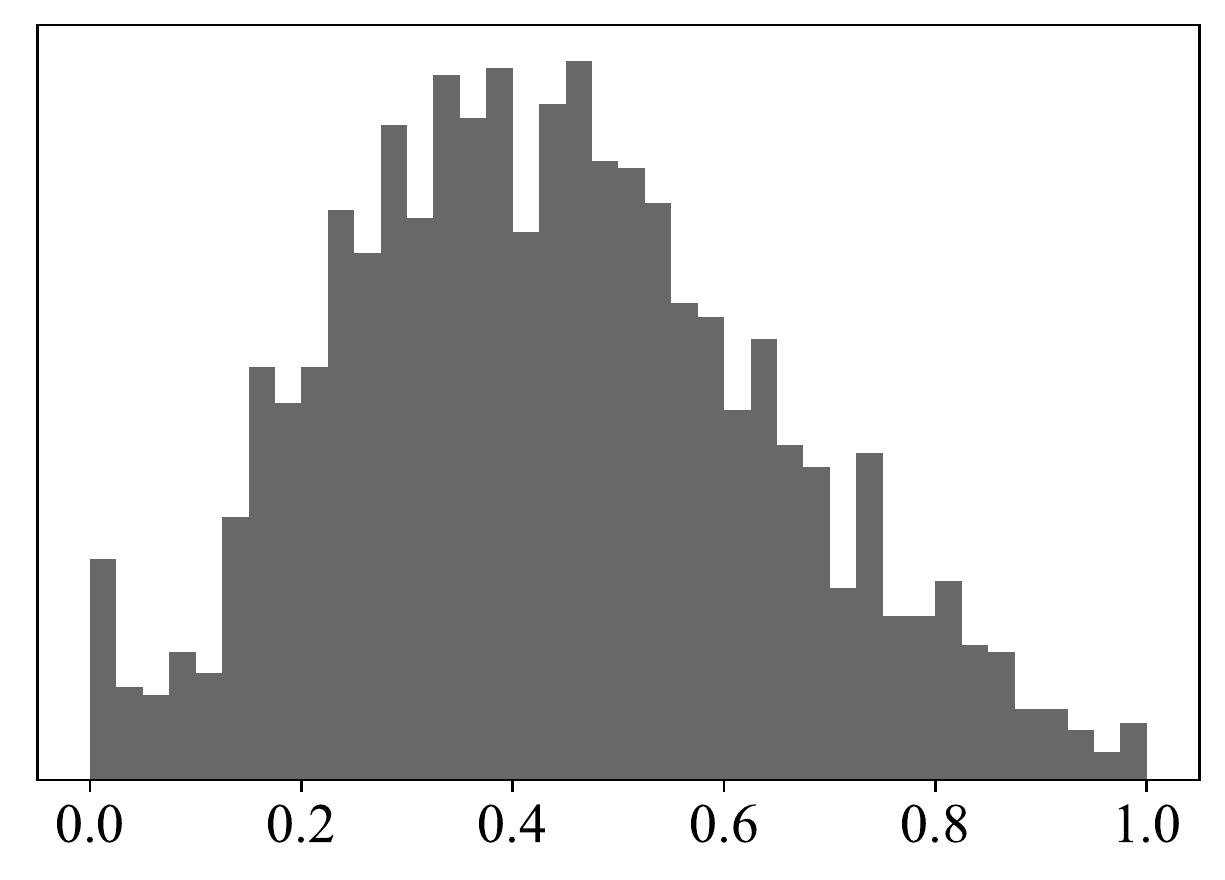}
        \caption{PTT $\times$ PTO}
    \end{subfigure}
    \caption{Distribution of cosine similaritiy values between ENO, PTT and PTO.}
    \label{fig:csdistributionslang}
\end{figure}

Average cosine similarity measures for each caption type pair – ENO x PTT, ENO x PTO and PTT x PTO – are presented in Table \ref{table:cscaptions}. The distributions of those similarities are shown in Figure \ref{fig:csdistributionslang}.

\begin{table}[!ht]
\centering
\def\arraystretch{1.4}
\begin{tabular}{lcccc}
\hline
    & \multicolumn{2}{c}{ENO} & \multicolumn{2}{c}{PTO} \\ \hline
    & avg. cos     & stdev    & avg. cos     & stdev    \\ \hline
ENO & -            & -        & 0.33         & 0.14     \\ \hline
PTT & 0.51         & 0.14     & 0.43         & 0.2      \\ \hline
PTO & 0.33         & 0.14     & -            & -        \\ \hline
\end{tabular}
\caption{Average cosine similarity between Associative\_arrays built for the frame semantic representations of ENO, PTT and PTO captions}{\label{table:cscaptions}}
\end{table}

Taking into consideration how the cosine similarity distributions shown in Figure \ref{fig:csdistributionslang} approximate a normal distribution and the almost equivalent variances, Student's and Welch's t-test were used to verify the significance of the differences, according to the variables variance. The cosine similarities between PTT and ENO (\textit{M} = 0.51, \textit{SD} = 0.14) were significantly higher than those between PTO and ENO (\textit{M} = 0.33, \textit{SD} = 0.14), $\textit{t}(1998) = 41.78, \textit{p} < 0.001$. Additionally, the average similarities between PTO and PTT (\textit{M} = 0.43, \textit{SD} = 0.2) are higher than those between PTO and ENO, with test statistic $\textit{t}(1998) = 19.98, \textit{p} < 0.001$. At the same time, those similarities are significantly smaller than those between PTT and ENO, with $\textit{t}(1998) = -13.71, \textit{p} < 0.001$.

\subsection{Similarity of semantic representations across modes in different annotation setups}

\begin{figure}[!ht]
    \begin{subfigure}[t]{0.48\linewidth}
        \centering
        \includegraphics[width=\linewidth]{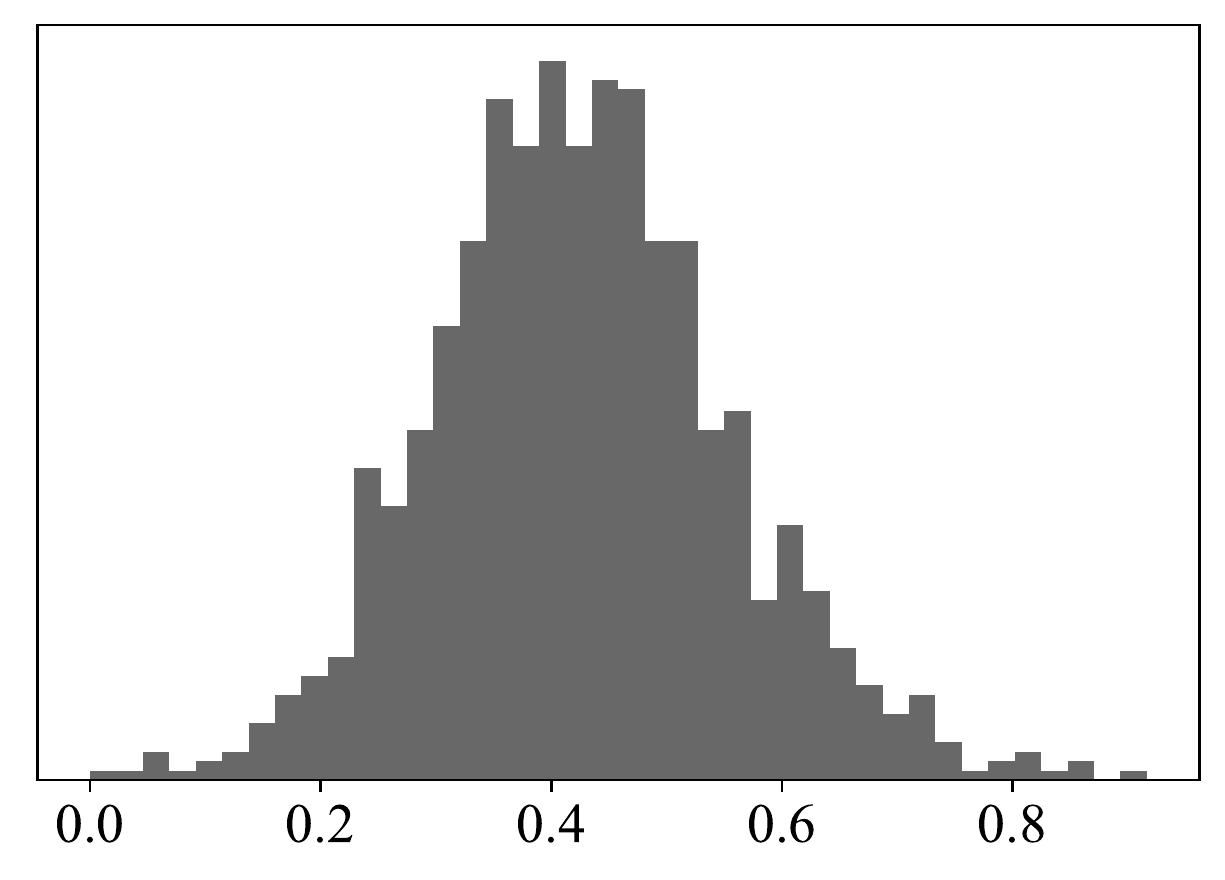}
        \caption{ENO $\times$ VWC}
    \end{subfigure}
    ~
    \begin{subfigure}[t]{0.48\linewidth}
        \centering
        \includegraphics[width=\linewidth]{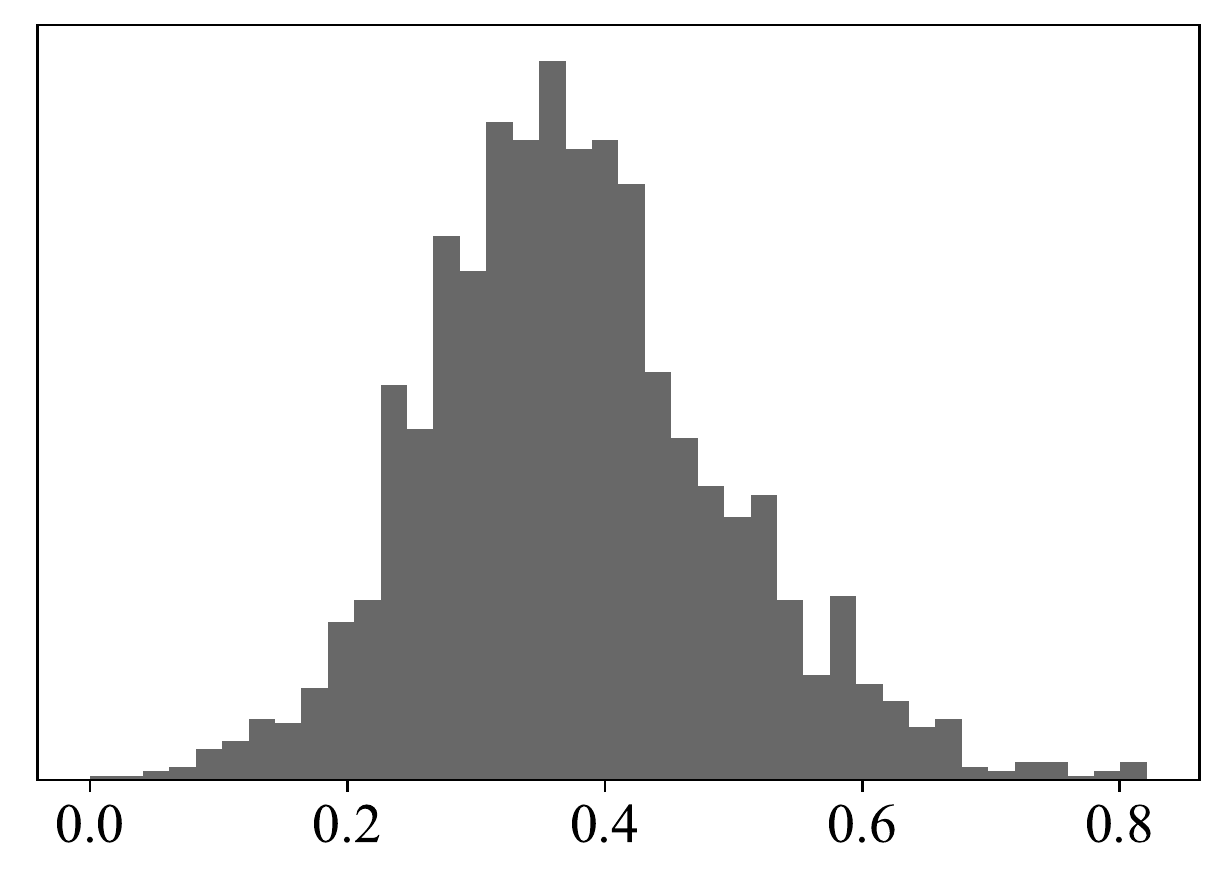}
        \caption{ENO $\times$ VWoC}
    \end{subfigure}
    \caption{Distribution of cosine similaritiy values between ENO, VWC and VWoC.}
    \label{fig:csdistributionsmodes}
\end{figure}

Average cosine similarity measures and the standard deviation for image annotations compared to the original English textual annotations are presented in Table \ref{table:similarity_annosetups}. The distributions of those similarities are shown in Figure \ref{fig:csdistributionsmodes}. The annotations made with a reference caption in English (VWC), when compared against ENO, had higher cosine similarities (\textit{M} = 0.43, \textit{SD} = 0.13) than the ones annotated without any reference (VWoC) (\textit{M} = 0.38, \textit{SD} = 0.12), with Student's t-test statistic $\textit{t}(998) = 8.64, \textit{p} < 0.001$.

\begin{table}[!ht]
\centering
\def\arraystretch{1.4}
\begin{tabular}{lcc}
\hline
      & \multicolumn{2}{c}{ENO} \\ \hline
      & avg. cos     & stdev    \\ \hline
VWC   & 0.43         & 0.13     \\ \hline
VWoC  & 0.38         & 0.12     \\ \hline
\end{tabular}
\caption{Similarity for image frame annotation setups with and without captions present.}
\label{table:similarity_annosetups}
\end{table}

\section{Discussion}
\label{sec:discussion}

The similarity coefficients from the multilingual comparison indicate that in terms of perspective of annotation, PTT seems to be somewhere between ENO and PTO, with a comparable distance to either. In some way, this was to be expected, given that the translation brief required translators to make a close rendition of the English original. With the cumulative analysis performed here, we cannot make the claim that this a clear case of \textit{shining through} \cite{teich_cross-linguistic_2003}, i.e. a case of source text features being over-represented in the target text.  For the domain of motion events, systematic framing differences between languages are well documented \cite{talmy_toward_2000-2,slobin_many_2004}. In the image annotation, however, we will find a vastly broader annotation including people and artifacts, and we do not yet know whether any of the involved languages has a preference for a certain framing or interpretation of these categories in image descriptions. Also, the variation we see between ENO and PTO might be due to randomness, with PTT being an intented close rendition of ENO. 

Besides a better understanding of potential framing preferences in a language, we also need to cross-check against variation that can happen within one language, or in other words: If we had two sets of annotations for English by different annotators, might they be as similar/distant as ENO/PTO or rather as ENO/PTT? While the numbers presented here may be first indicators, clearly more elaborate setups and evaluations are needed to dig deeper into the questions of framing preferences between languages and translation effects. As to the latter, it is not a given that influence from the source language is the only factor to be taken into account: As shown, e.g., in \cite{vandevoorde_semantic_2020}, \textit{normalization} \cite{baker_corpora_1995}, i.e. the (over-)adherence to target language conventions is another semantic effect that can be witnessed in translation, or re-framings due to an open list of factors such as those described, i.a., in \cite{slobin_relating_2005,rojo_constructing_2013,czulo_aspects_2017,ohara-2020-finding}.

As for the image annotation setups, the experiments indicate that, under a model that takes perspective into account, labels assigned to images are not to be taken as some sort of ``ground truth'' representation. Annotators taking part in the experiment tended to frame the image according to the caption, which is shown by a significant higher cosine similarity between ENO and VWC. The very examples in Figures \ref{fig:caption-anno} and \ref{fig:vis_anno} give an indication of that: while the person holding the suitcase was annotated as a thief when in presence of the caption, they were tagged as a person in the absence of the caption clue.

This is an indication that the role of images as proxies for ``ground truth'' in multimodal datasets is, at the very least, limited, if one considers that meanings are relativized to perspectivized scenes, as pointed out by \newcite{fillmore1977case}. The main issue at stake here is that, in general, image annotation in multimodal datasets involve the assignment of labels from a categorization system that does not encode perspective.

In cases where the annotation is carried out by humans, which is the case for the Flickr 30k Entities dataset, the categories available for annotation of the bounding boxes are very coarse-grained and include only: people, clothing, body parts, animals, vehicles, instruments, scene, not visual and other. In this scenario, the distinction between thief and person, for instance, would be subsumed under the ``people'' tag. The scenario is even more concerning when we consider that, among the 559,767 bounding boxes in the Flickr 30k Entities dataset, 182,136 (32.53\%) received the ``people'' tag and 138,658 (24.77\%) received the ``other'' tag.

In the cases where image annotation is conducted automatically, using computer vision algorithms such as YOLOv3 \cite{redmon2018yolov3}, for example, the core problem remains unchanged. Such systems are trained on datasets such as MS-COCO \cite{10.1007/978-3-319-10602-1_48} or Open Images \cite{OpenImages}. In both cases, the categories used for tagging bounding boxes are organized in ontologies that do not encode perspective either. Last but not least, even in the case of Open Images, where, on top of the categories assigned to each image, attributes and relations can also be assigned as triplets, it is not made clear whether the different perspectives on them are encoded.

Because the way humans interpret the entities in an image may be influenced by the text accompanying it – as the results in Table \ref{table:similarity_annosetups} suggest –, the current configuration of image annotation systems may limit the role of images in downstream tasks such as multimodal machine translation. Even in Flickr 30k, where captions are conceptual descriptions of images, the relation between image and text is not equivalent to that of an absolute fact and one possible description of it. Images may too accommodate different perspectives and accounting for those differences is key for assuring that the relation between the image and the text is preserved, for instance, in a translated sentence.   

\section{Conclusion(s) and further work}
\label{sec:conclusion}

The experiments and analyses described in this contribution have produced two results: 
\begin{enumerate}
    \item Frame semantic similarity for image captions in different languages are sensitive to whether a description is a translation or not.
    \item Semantic similarity was also influenced by the annotation setup in that presenting captions with images to be annotated produced higher semantic similarity across modes, indicating that image description data cannot be assumed to be an ``independent source'' or ``ground truth'' of semantic information.
\end{enumerate}
The former result needs further investigation in order to test for its generalizability. This would include extending the experiment to different language pairs as well as comparing such results with variation within a language, e.g., with multiple captions per image in one language. Also, the question of whether there are language-specific framing preferences at play can only be answered by means of deeper analyses of the frame semantic annotation of the original captions.

The latter result in a sense is analogous to the observation that the translation of image captions seems to carry, in parts, influence from an ``anterior'' - for the case of translation, this the description to be translated, with regard annotation setup, shown to annotators. If priming is to be assumed as a relevant factor in both cases, then the observation does not come as a surprise, but at the same time commands caution for future annotation setups. Priming as a factor could be tested for in order to corroborate this assumption.

\section{Acknowledgements}

Authors acknowledge the support of the Graduate Program in Linguistics at the Federal University of Juiz de Fora. Research presented in this paper is funded by CAPES PROBRAL grant 88887.144043/2017-00 and CNPq grants 408269/2021-9 and 315749/2021-0. Viridiano's research was funded by CAPES PROBRAL PhD exchange grant 88887.628830/2021-00.

\section{Bibliographical References}\label{reference}

\bibliographystyle{lrec2022-bib}
\bibliography{lrec2022-example}


\end{document}